 \documentclass[pmlr,twocolumn,10pt]{jmlr} % W&CP article

\usepackage{booktabs}
\usepackage{siunitx}

\usepackage[switch]{lineno}

\usepackage{listings}
\usepackage{color} % Optional, for colored code highlighting
\usepackage{cite}
\usepackage{amsmath,amssymb,amsfonts}
\usepackage{algorithmic}
\usepackage{graphicx}
\usepackage{textcomp}
\usepackage{xcolor}
\usepackage{booktabs}
\usepackage{float}
\usepackage{adjustbox}
\usepackage{subcaption}
\usepackage{subcaption}

\theorembodyfont{\upshape}
\theoremheaderfont{\scshape}
\theorempostheader{:}
\theoremsep{\newline}

\jmlrvolume{XXX}
\jmlryear{2025}
\jmlrworkshop{Machine Learning for Health (ML4H) 2025} % W&CP title

 \title[Short Title]{ML4H 2025 Template: Proceedings Track}

 \title{Longitudinal Progression Prediction of Alzheimer's Disease with Tabular Foundation Model}

\author{%
\Name{Yilang Ding}$^{1}$%
\Email{lance.ding@emory.edu}\\
\Name{Jiawen Ren}$^{1}$%
\Email{jren46@emory.edu}\\
\Name{Jiaying Lu}$^{2}$%
\Email{jiaying.lu@emory.edu}\\
\Name{Hyunjung Gloria Kwak}$^{2}$%
\Email{hyunjung.kwak@emory.edu}\\
\Name{Armin Iraji}$^{3,4}$%
\Email{airaji@gsu.edu}\\
\Name{Shengpu Tang}$^{1}$%
\Email{shengpu.tang@emory.edu}\\
\Name{Alex Fedorov}$^{2}$%
\Email{avfedor@emory.edu}\\
\addr $^{1}$Department of Computer Science, Emory University, Atlanta, GA, USA\\
\addr $^{2}$Center for Data Science, Nell Hodgson Woodruff School of Nursing, Emory University, Atlanta, GA, USA\\
\addr $^{3}$Department of Computer Science, Georgia State University, Atlanta, GA, USA\\
\addr $^{4}$Tri-Institutional Center for Translational Research in Neuroimaging and Data Science (TReNDS), Atlanta, GA, USA
}

\begin{document}

\maketitle

\begin{abstract}
Alzheimer's disease is a progressive neurodegenerative disorder that remains challenging to predict due to its multifactorial etiology and the complexity of multimodal clinical data. Accurate forecasting of clinically relevant biomarkers, including diagnostic and quantitative measures, is essential for effective monitoring of disease progression. This work introduces L2C-TabPFN, a method that integrates a longitudinal-to-cross-sectional (L2C) transformation with a pre-trained Tabular Foundation Model (TabPFN) to predict Alzheimer's disease outcomes using the TADPOLE dataset. L2C-TabPFN converts sequential patient records into fixed-length feature vectors, enabling robust prediction of diagnosis, cognitive scores, and ventricular volume. Experimental results demonstrate that, while L2C-TabPFN achieves competitive performance on diagnostic and cognitive outcomes, it provides state-of-the-art results in ventricular volume prediction. This key imaging biomarker reflects neurodegeneration and progression in Alzheimer's disease. These findings highlight the potential of tabular foundational models for advancing longitudinal prediction of clinically relevant imaging markers in Alzheimer's disease.
\end{abstract}

\paragraph*{Data and Code Availability}
The benchmarking datasets used in this study are derived from the Alzheimer’s Disease Neuroimaging Initiative (ADNI) database (\url{adni.loni.usc.edu}). ADNI provides a multimodal longitudinal dataset that includes imaging (MRI, PET), clinical assessments, and neuropsychological evaluations, enabling prediction of Alzheimer’s disease progression. The ADNI project was launched in 2003 as a public–private partnership and is led by Principal Investigator Michael W. Weiner, MD.

The code will be made publicly available upon publication.

\paragraph*{Institutional Review Board (IRB)}
This study involved secondary analysis of publicly available, de-identified data and did not require IRB approval.

\section{Introduction}
Alzheimer's disease imposes a rapidly escalating global health burden. Currently, approximately 7.2 million Americans aged 65 or older are living with Alzheimer's dementia, a number expected to nearly double by 2050~\citep{alzheimer2025}. Alzheimer's disease is a progressive neurodegenerative disorder marked by a gradual decline in memory and cognitive function, representing the most common cause of dementia~\citep{alzheimer2025}. Its multifactorial etiology involves complex genetic, biochemical, and environmental interactions that have posed significant challenges in developing effective treatments~\citep{zhang2024recent}.

To guide clinical trials and improve patient care, researchers have turned to computational models of disease progression~\citep {young2024data}. These frameworks require longitudinal cohorts, including clinical assessments, neuroimaging metrics, cognitive scores, and fluid biomarkers, to chart individual trajectories over time. By capturing baseline-to-follow-up changes, they help identify early predictors of decline, quantify progression rates, and characterize the progression trajectory in affected individuals.

A landmark in this field is the TADPOLE Challenge~\citep{marinescu2019tadpole}, which provides a standardized, preprocessed multimodal dataset drawn from the Alzheimer's Disease Neuroimaging Initiative (ADNI). It contains predictions for three key outcomes: diagnostic status, ADAS-Cog scores, and ventricular volume. TADPOLE has catalyzed the development and benchmarking of predictive methods.

Traditional approaches, such as the Frog model, a framework with longitudinal-to-cross-sectional transformation (L2C) and XGBoost, have demonstrated robust performance in the TADPOLE challenge~\citep{marinescu2020alzheimer,zhang2024cross}. Frog leverages an L2C transformation that converts sequential patient records into fixed-length feature vectors, enabling XGBoost~\citep{chen2016xgboost} to capture temporal trends efficiently. Despite the emergence of neural network architectures~\citep{zhang2024cross}, XGBoost~\citep{chen2016xgboost} remains a strong baseline because of its efficiency and interpretability for tabular data. Moreover, neural networks have historically struggled with tabular datasets due to overfitting and inadequate handling of structured data. The advent of Tabular Foundation Models (TabPFN)~\citep{hollmann2023tabpfn,hollmann2025tabpfn} has shifted this paradigm. TabPFN leverages pre-training on a synthetic tabular data, which later allows to generalize better on small datasets via in-context learning. It compete with or surpass traditional machine learning approaches~\citep{hollmann2025tabpfn}.

Motivated by these advancements, we propose harnessing the potential of TabPFN to improve predictions of longitudinal Alzheimer's disease progression. Our contributions are as follows. We present the first application of a tabular foundation model (TabPFN) with longitudinal-to-cross-sectional (L2C) transformation for longitudinal forecasting in Alzheimer's disease. To address irregular sampling, missing events, and bias from varying patient entry points, we use a longitudinal-to-cross-sectional (L2C) transformation with engineered features and a horizon variable that reflects relative prediction intervals. This enables the use of traditional classification and regression methods on longitudinal data. We benchmark our approach on the TADPOLE dataset against the XGBoost-based Frog model, showing that while both models perform well on diagnosis and cognitive scores, L2C-TabPFN achieves better results in ventricular volume prediction. We also provide interpretability analysis and a reproducible baseline for future work.

\section{Methods}
In this study, we introduce L2C-TabPFN model~\citep{hollmann2025tabpfn,hollmann2023tabpfn} and evaluate it on the TADPOLE dataset~\citep{marinescu2019tadpole}. Then, we compare it with the state-of-the-art (SOTA) baseline Frog~\citep{marinescu2020alzheimer}.

\subsection{Dataset: TADPOLE}

In this work, we employ three tasks to evaluate the capability of machine learning models in predicting the progression of Alzheimer's disease (AD) using longitudinal data:

\begin{itemize}
\item \textbf{DX}: Predicting the clinical diagnosis across three stages of AD progression—cognitively normal, mild cognitive impairment (MCI), and probable AD.
\item \textbf{ADAS-Cog}: Predicting scores on the AD Assessment Scale–Cognitive Subscale, which measures cognitive function.
\item \textbf{Ventricle}: Predicting the volume of brain ventricles, normalized by intra-cranial volume, as an indicator of brain atrophy.
\end{itemize}

The benchmarking datasets are adapted from The Alzheimer's Disease Prediction Of Longitudinal Evolution (TADPOLE) Challenge~\citep{marinescu2019tadpole}.
The challenge uses data from the Alzheimer's Disease Neuroimaging Initiative (ADNI) database (\url{adni.loni.usc.edu}).
The ADNI database offers a multimodal longitudinal dataset with imaging (MRI, PET) and clinical and neuropsychological assessments to predict the progression of Alzheimer's disease.
The ADNI project was led by Principal Investigator Michael W. Weiner, MD, and started in 2003 as a public-private collaboration.

The TADPOLE Challenge dataset contains four preprocessed data sets: \textbf{D1}, \textbf{D2}, \textbf{D3}, and \textbf{D4}. \textbf{D1} is a training set that captures longitudinal data across three phases of ADNI: ADNI1, ADNI GO, and ADNI2.
For every participant, there are at least two separate visits. \textbf{D2} contains data from the rollover individuals and is used as a prediction and forecasting dataset.
\textbf{D3} is a cross-sectional dataset containing single time point observations.
\textbf{D4} is also from rollover participants, which contain at least one of the three outcomes.
We used only \textbf{D1} and \textbf{D2} in our experiments.

The challenge organizers already preprocessed the dataset into a tabular format:
\begin{itemize}
    \item \textbf{Demographics:} This category includes APOE4 genotype status, sex ($\text{is\_male}$), education level (educ), marital status, current age, and the number of months since baseline ($\text{month\_since\_baseline}$).
    \item \textbf{Clinical diagnosis:} Categorical labels indicating the participant's diagnostic group: cognitively normal (CN), mild cognitive impairment (MCI), or Alzheimer's disease (AD).
    \item \textbf{Clinical assessments:} Numeric or ordinal scores that measure various aspects of cognitive and functional status. For instance, the Mini-Mental State Examination (MMSE) ranges from 0 to 30, and the Clinical Dementia Rating (CDR) Global Score ranges from 0 to 3.
    \item \textbf{Imaging biomarkers:} Quantitative brain volume measurements from magnetic resonance imaging (MRI), such as ventricle volume, fusiform volume, whole-brain volume, hippocampus volume, middle temporal volume, and intracranial volume (ICV). These continuous measures are essential for tracking structural brain changes associated with Alzheimer's disease progression.

\end{itemize}

\begin{figure}[t]
    \centering
    \includegraphics[width=0.9\linewidth]{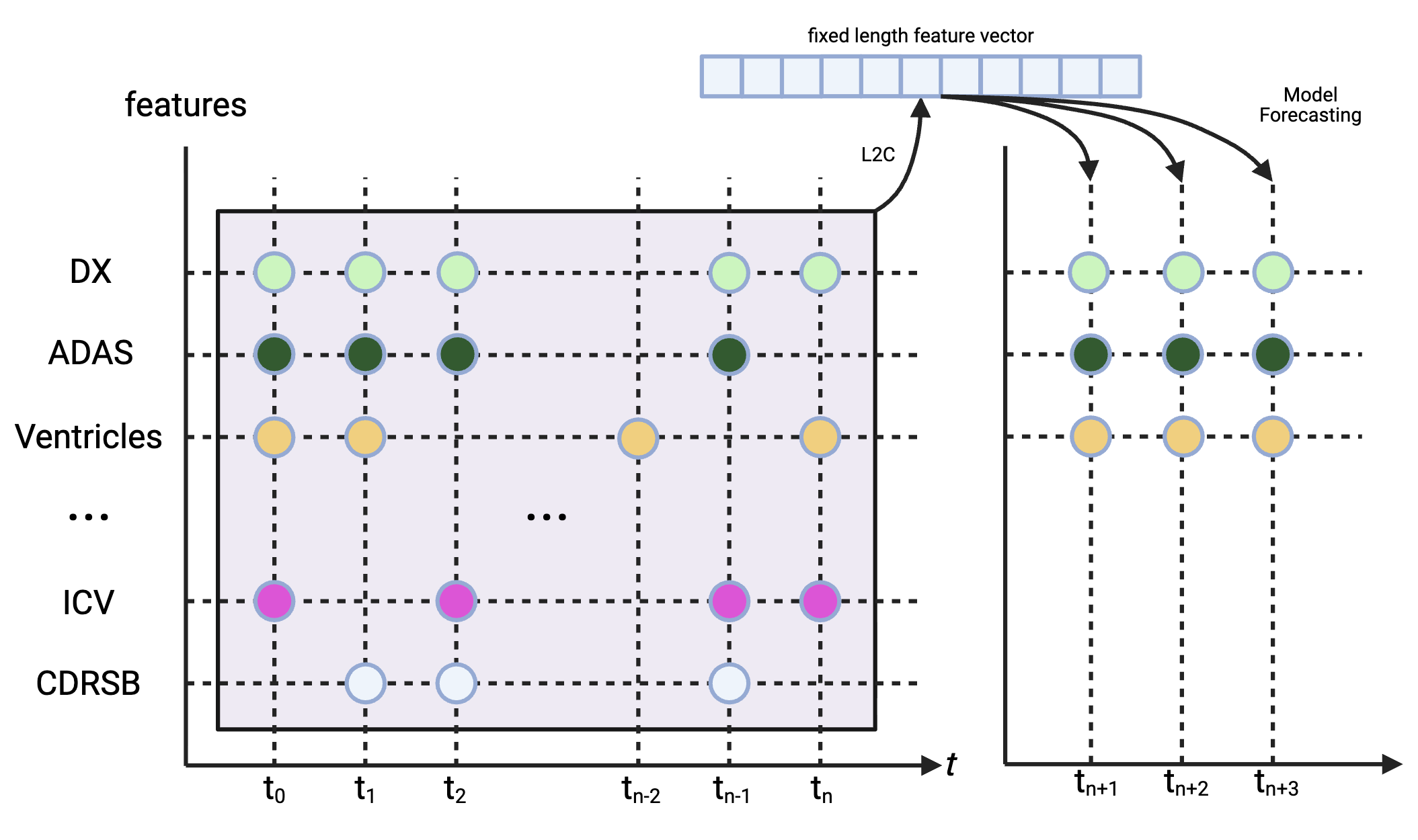}
    \caption{Illustration of the longitudinal-to-cross-sectional (L2C) feature transformation and forecasting process.
    Longitudinal multimodal features such as diagnosis (DX), ADAS scores, ventricular volumes, intracranial volume (ICV),
    and CDRSB are organized over multiple timepoints. The L2C transformation converts these into a fixed-length feature
    vector, which serves as input to the forecasting model. The model then predicts clinical and imaging outcomes for
    future timepoints.}
    \label{fig:l2c-feature}
    \vspace{-1em}
\end{figure}

\subsection{Longitudinal-to-Cross-sectional (L2C) Transformation}

The L2C transformation~\citep{zhang2024cross,marinescu2020alzheimer} converts each patient's sequential records into fixed-length feature vectors by summarizing historical data up to (but not including) the current time point. For each patient, observations are grouped and sorted by baseline month. Then, for each observation at time \(t\) (starting from the second record), all previous observations (i.e., for \(s < t\)) are aggregated to compute summary statistics for each numeric feature \(x\) (e.g., MMSE, CDRSB, Ventricles).

Specifically, for a given feature \(x\), we compute:
\begin{itemize}
    \item \textbf{Most recent measurement:}
    \begin{equation}
        mr_x = x(t^*), \quad \text{where } t^* = \max\{s \mid s < t\}
    \end{equation}

    \item \textbf{Time since most recent measurement:}
    \begin{equation}
        \Delta t_{mr} = t - t^*
    \end{equation}

    \item \textbf{Change rate:}

    \begin{align}
        mr\_change\_x &= \frac{x(t^*) - x(t^{**})}{t^* - t^{**}} \nonumber\\
        &\text{where } t^{**} = \max\{s \mid s < t^*\}
    \end{align}

    \item \textbf{Minimum value:}
    \begin{equation}
        low_x = \min\{x(s) \mid s < t\}
    \end{equation}
    \begin{equation}
        \Delta t_{low} = t - \min\{s \mid x(s) = low_x\}
    \end{equation}

    \item \textbf{Maximum value:}
    \begin{equation}
        high_x = \max\{x(s) \mid s < t\}
    \end{equation}
    \begin{equation}
        \Delta t_{high} = t - \min\{s \mid x(s) = high_x\}
    \end{equation}
\end{itemize}
For the diagnostic variable (DX), additional indicators (e.g., the most recent diagnosis, a "milder" state flag, and the time since the best or worst diagnosis) are computed to capture disease progression trends.

Finally, demographic variables and outcome targets (such as DX, ADAS-Cog, and Ventricles) are appended for the current time point. This process yields multiple cross-sectional snapshots per patient.

\subsection{Augmentation}
To generate more training data, we apply the same augmentation technique as proposed by authors of the related work~\citep{zhang2024cross}. For a patient with $n$ data points with the time of the first visit denoted as $t_1$, for all $t_m:1\le m<n$ we run L2C up to $t_m$, and use it to predict $t_{m+1}, t_{m+2},...,t_n$. This generates a total of $\frac{n(n-1)}{2}$ distinct training examples. To account for this change, as well as the fact that the time between visits for each patient is not constant, we augment the `month\_bl` feature into the `horizon` feature, which encodes the prediction horizon from the time point that we stop L2C on to the time point that we want to predict. After these steps, the number of training examples for each task is shown in Table~\ref{tab:dataset-split-numbers}.

\subsection{Tabular Foundation Model (TabPFN)}
We propose to use the Tabular Foundation Model (TabPFN), a transformer‐based architecture tailored for small to medium-sized tabular datasets (up to 10,000 samples and 500 features). In our implementation, we first apply the L2C transformation (described above) to convert each patient's longitudinal records into fixed-length cross-sectional representations (see Figure~\ref{fig:l2c-feature}). These transformed snapshots are then used as input to TabPFN, which leverages in-context learning (ICL) to process the entire dataset in a single forward pass. Pre-trained on millions of synthetic datasets generated via a structural causal model (SCM), TabPFN is exposed to diverse data challenges, including missing values, heterogeneous feature scales, and outliers, and thereby learns a robust prediction algorithm. Importantly, once pre-trained, TabPFN is applied directly to new, real-world datasets without additional training.

Mathematically, given a training set \((X_{\text{tr}}, y_{\text{tr}})\) and a test set \(X_{\text{test}}\) (obtained after the L2C transformation), TabPFN approximates the posterior predictive distribution:
\begin{equation}
p(y_{\text{test}} \mid X_{\text{tr}}, y_{\text{tr}}, X_{\text{t}}) \approx \text{TabPFN}(X_{\text{tr}}, y_{\text{tr}}, X_{\text{t}}).
\end{equation}
The pre-training objective was to minimize the negative log-likelihood:
\begin{equation}
\mathcal{L} = -\mathbb{E}\left[\log p(y_{\text{t}} \mid X_{\text{tr}}, y_{\text{tr}}, X_{\text{t}})\right].
\end{equation}
TabPFN's architecture adapts the standard transformer encoder to the two-dimensional structure of tabular data by assigning a distinct embedding to each cell. It employs a two-way attention mechanism: first, inter-feature attention aggregates information within each sample (row), and then inter-sample attention aggregates information across samples (column). To avoid redundant computations, the keys and values for training samples are cached and reused during inference, enabling efficient prediction in a single forward pass.

\begin{figure}[t]
    \centering
    \includegraphics[width=\linewidth]{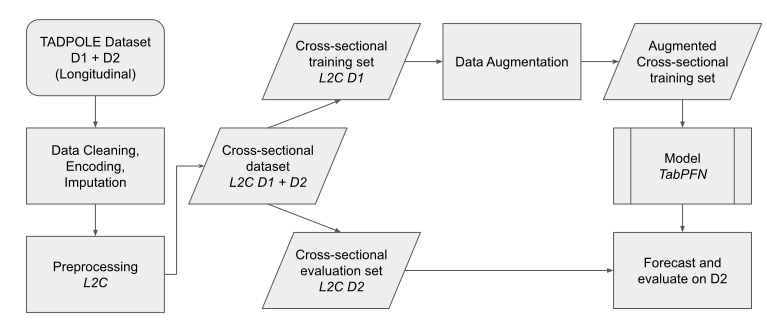}
    \caption{Overview of the experimental pipeline for L2C-TabPFN forecasting. The process starts with the TADPOLE dataset
    and proceeds through data cleaning, preprocessing, and longitudinal-to-cross-sectional (L2C) feature transformation.
    The pipeline then creates cross-sectional training and evaluation sets, applies data augmentation to the training set,
    and trains the TabPFN model. Model performance is evaluated on the holdout set D2.}
    \label{fig:l2c-tabpfn-flow}
\end{figure}

\begin{table}[t]
    \centering
    \caption{Dataset splits for each task after preprocessing. Numbers for cross-validation are averaged and rounded to the nearest integer. Note that to prevent overfitting on patients from the training set to the validation set, we separated the dataset into approximately equal folds by patient, ensuring that no patient was included in both the training set and the validation set simultaneously. For validation, we used the forecasting process, using the first half of available patient information to predict the second half. As such, only training data that represented exactly the first half of the patient history were used for validation, which caused the apparent small size of the evaluation set.}
    \footnotesize
    \begin{tabular}{lccccc}
        \hline
        \textbf{Task} & \textbf{Before augmentation} & \textbf{Total Train Size} & \textbf{CV Train Size} & \textbf{Validation Size} & \textbf{Test Size} \\
        \hline
        DX          & 3,677 & 7,340 & 5,872 & 247 & 2,538 \\
        ADAS-Cog    & 3,591 & 7,075 & 5,660 & 242 & 2,551 \\
        Ventricles  & 3,256 & 4,990 & 3,992 & 183 & 1,241 \\
        \hline
    \end{tabular}
    \label{tab:dataset-split-numbers}
\end{table}

\subsection{Experimental Details}
Our experimental pipeline was implemented in Python (Scikit-learn~\citep{pedregosa2011scikit}). The pipeline comprised several steps: beginning with data preprocessing, feature transformation using the L2C pipeline, model training, and finally evaluation for both classification (DX) and regression tasks (ADAS-Cog, Ventricles), as illustrated in Figure~\ref{fig:l2c-tabpfn-flow}.

% FLOWCHART HERE
% https://docs.google.com/presentation/d/1bjeflqNUhZ8v3wVeQSruEcnLO7c04p6sjW80Jj2SmYI/edit?usp=sharing

We first preprocessed the TADPOLE dataset by sorting all patient records by months since baseline and re-encoded categorical variables such as sex, marital status, age, and education. The longitudinal-to-cross-sectional (L2C) transformation was then applied to the longitudinal features, including clinical diagnosis, MMSE score, CDR score, and brain region volumes. For each patient, sequential records were grouped by baseline month, and summary statistics (e.g., most recent measurement, change rate, minimum and maximum values, along with corresponding time intervals) were computed for each numeric feature. Additional indicators for the diagnostic variable (DX) were also derived to capture trends in disease progression. The transformation produced a fixed-length feature vector, along with demographic information and outcome targets (DX, ADAS-Cog, and Ventricles).

In the classification task for model training, patients designated as D1 (training set) were used to build predictive models, while patients marked as D2 served as the testing set. We tuned hyperparameters using Optuna~\citep{akiba2019optuna} with cross-validation (StratifiedKFold for classification and KFold for regression) over 90 trials. The TabPFN hyperparameter search ranges for all tasks are shown in Table~\ref{tab:hyperparam-search}:

\begin{table}[h]
    \centering\footnotesize
    \caption{TabPFN Optuna Hyperparameter Search Space}
    \begin{tabular}{lcc}
        \hline
        \textbf{Hyperparameter} & \textbf{Search Space} & \textbf{Type} \\
        \hline
        \texttt{n\_estimators} & 1 to 31 (step size 2) & Integer \\
        \texttt{softmax\_temperature} & 0.1 to 2.0 (log-uniform) & Continuous \\
        \texttt{average\_before\_softmax} & \{True, False\} & Categorical \\
        \hline
    \end{tabular}
    \label{tab:hyperparam-search}
\end{table}

The hyperparameters that were produced and used in inference are shown in Table~\ref{tab:final-hyperparams}.

\begin{table}[h]
\footnotesize
    \centering
    \caption{TabPFN Final Hyperparameters}
    \begin{tabular}{l c c c}
        \hline
        \textbf{Hyperparameter} & \textbf{Ventricles} & \textbf{ADAS} & \textbf{DX} \\
        \hline
        $n\_estimators$ & 31 & 9 & 25 \\
        $softmax\_temperature$ & 0.718 & 1.212 & 1.981 \\
        $average\_before\_softmax$ & True & True & True \\
        \hline
    \end{tabular}

    \label{tab:final-hyperparams}
\end{table}

Similarly, for XGB, we utilized 100 Optuna trials per model to select hyperparameters. The search range follows the recommendation of~\citep{zhang2024cross} which shown in Table~\ref{tab:hyperparam-search-xgb}.

\begin{table}[h]
\footnotesize
    \centering
    \caption{XGBoost Optuna Hyperparameter Search Space}
    \begin{tabular}{lcc}
        \hline
        \textbf{Hyperparameter} & \textbf{Search Space} & \textbf{Type} \\
        \hline
        \texttt{max\_depth} & 3 to 8 & Integer \\
        \texttt{subsample} & 0.4 to 1.0 & Continuous \\
        \texttt{learning\_rate} & 0.01 to 0.2 & Continuous \\
        \texttt{n\_estimators} & 100 to 1000 (step size 50) & Integer \\
        \hline
    \end{tabular}

    \label{tab:hyperparam-search-xgb}
\end{table}

Resulting XGB hyperparameters are shown in Table~\ref{tab:final-hyperparam-search-xgb}.

\begin{table}[h]
\footnotesize
    \centering
    \caption{XGBoost Final Hyperparameters}
    \begin{tabular}{l c c c}
        \hline
        \textbf{Hyperparameter} & \textbf{Ventricles} & \textbf{ADAS} & \textbf{DX} \\
        \hline
        $max\_depth$ & 3 & 4 & 3 \\
        $subsample$ & 0.5826 & 0.5464 & 0.4618 \\
        $learning\_rate$ & 0.0149 & 0.0138 & 0.0102 \\
        $n\_estimators$ & 650 & 500 & 850 \\
        \hline
    \end{tabular}

    \label{tab:final-hyperparam-search-xgb}
\end{table}

For classification (predicting disease status), we evaluated performance using the multiclass area under the receiver-operating characteristic curve (AUROC) and balanced classification accuracy (BCA). For regression tasks (ADAS-Cog, Ventricles), we used mean absolute error (MAE) as the evaluation metric. After cross-validation, the best models were selected, and their performance metrics (mean and standard deviation across folds) were computed on the holdout test set D2. Frog~\citep{marinescu2020alzheimer} served as our state-of-the-art baseline and is built upon an XGBoost framework with L2C transform.

The application of L2C transformed the original data from variable-length patient histories to fixed-length feature vectors per time point, where the model is asked to predict the outcome given the L2C statistics up to a particular time point. To produce forecasts, we employ the L2C process to generate summary statistics up to the cutoff of available data. We then sweep the prediction horizon, producing an augmented L2C vector for each desired prediction horizon, with the relevant time-based statistics updated. In practice, this means incrementing features like 'horizon' and `time\_since\_mr\_XXX` by the required amount, while leaving the other L2C features that are not directly influenced by the horizon unchanged. This is analogous to asking the model to predict a specified horizon utilizing available ground truth information, which is captured in the L2C.

\begin{table*}[t]
    \centering
    \caption{Comparison of TabPFN and XGB results for forecasting. Models were trained on dataset D1 and used to forecast outcomes on dataset D2, with performance metrics averaged over five random seeds. Arrow directions indicate the desired improvement direction: higher is better for MAUC and BCA, and lower is better for MAE. Statistical significance was assessed using a Wilcoxon signed-rank test based on the \textbf{best} model in each task.}
    \label{tab:comparison2}
  \begin{tabular}{c|c|c|c|c|c}
\toprule
Task & Metric & L2C-TabPFN (Mean ± Std) & Frog (Mean ± Std) & p-value & Significance \\
\midrule
DX & MAUC $\uparrow$ & 0.9138 ± 0.0007 & \textbf{0.9258 ± 0.0003} & 0.0312 & * \\
DX & BCA $\uparrow$ & 0.7839 ± 0.0028 & \textbf{0.8056 ± 0.0019} & 0.0312 & * \\
ADAS-Cog & MAE $\downarrow$ & 6.7932 ± 0.0473 & \textbf{6.0030 ± 0.0229} & 0.0312 & * \\
Ventricles & MAE $\downarrow$ & \textbf{0.1577 ± 0.0005} & 0.1781 ± 0.0008 & 0.0312 & * \\
\bottomrule
\end{tabular}    \vspace{-1em}
\end{table*}

\begin{figure*}[ht]
    \centering
    \includegraphics[width=1\linewidth]{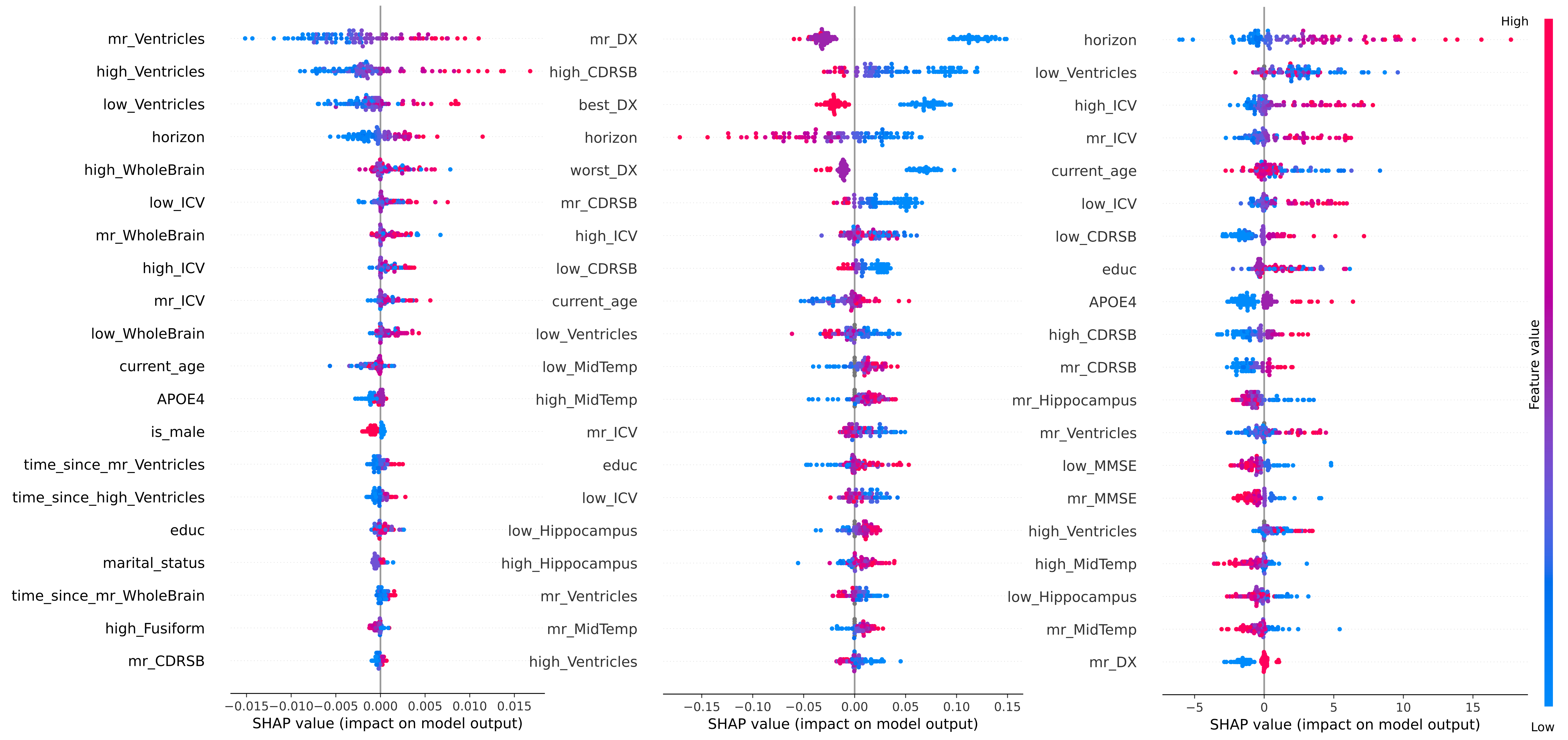}
    \caption{SHAP summary plots for feature importance in the ventricles, diagnosis (DX), and ADAS models for \textbf{L2C-TabPFN}. Each point represents the SHAP value for a specific feature in a single sample. Feature importance is ranked from top to bottom, with color indicating the feature value (blue for low and pink for high). The plots highlight which features have the greatest impact on model predictions across all samples.}
    \label{fig:shap_summary}
    \vspace{-1em}
\end{figure*}

\begin{figure*}[ht]
    \centering
    \includegraphics[width=1\linewidth]{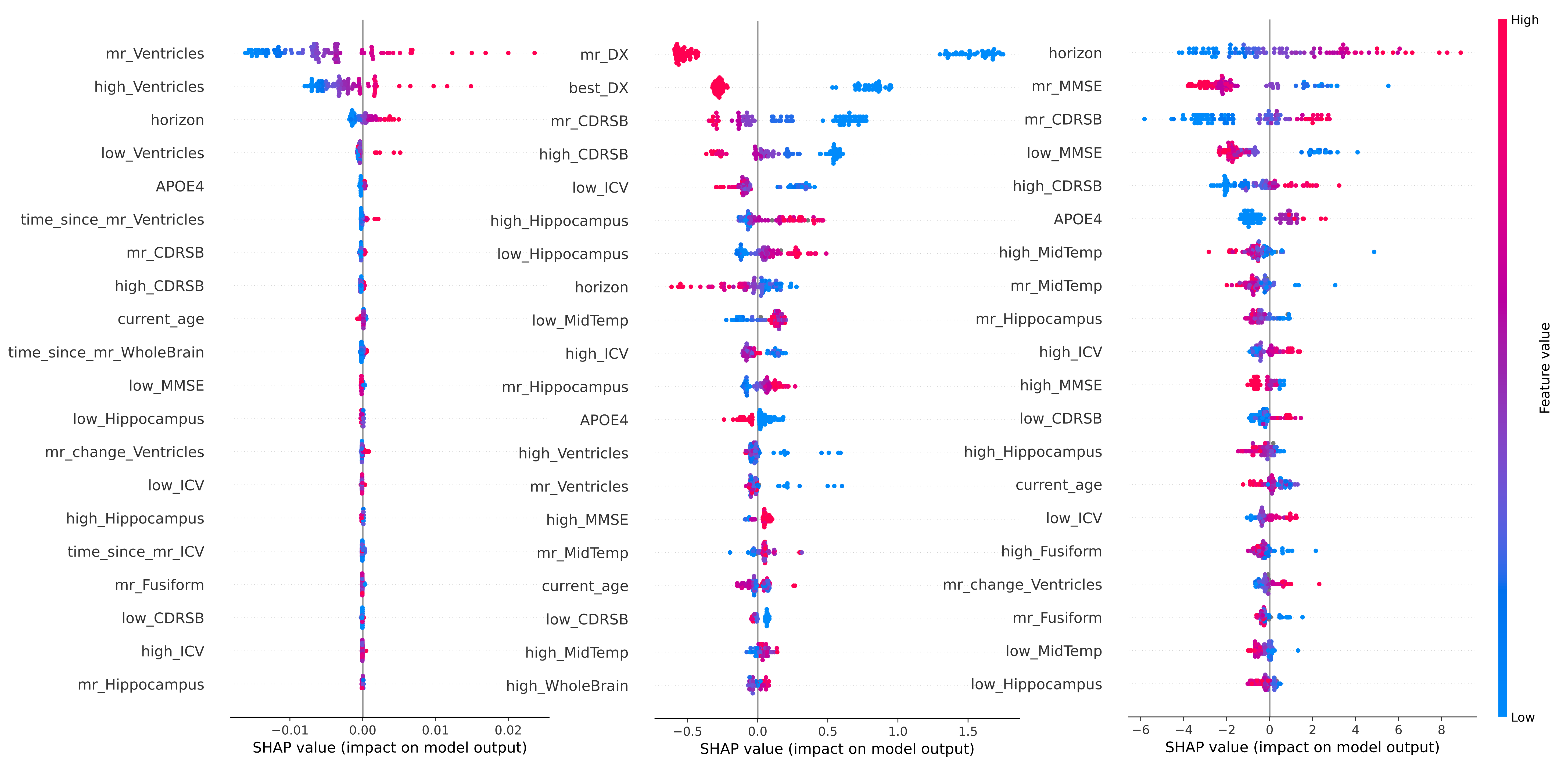}
    \caption{SHAP summary plots for feature importance in the ventricles, diagnosis (DX), and ADAS models for \textbf{Frog}. Each point represents the SHAP value for a specific feature in a single sample. Feature importance is ranked from top to bottom, with color indicating the feature value (blue for low and pink for high). The plots highlight which features have the greatest impact on model predictions across all samples.}
    \label{fig:shap_summary_xgb}
        \vspace{-1em}
\end{figure*}

\section{Results}

\subsection{Results on Forecasting}

Table~\ref{tab:comparison2} presents a comparison of L2C-TabPFN and Frog across three key forecasting tasks in continuous outcome prediction. The results highlight that the relative strengths of the two models depend on the target metric.

\textbf{Diagnostic classification (DX):} For both the MAUC (multi-class area under the curve) and BCA (balanced classification accuracy) metrics on the diagnostic (DX) task, Frog consistently outperforms L2C-TabPFN. Specifically, Frog achieves a higher MAUC of $0.9258 \pm 0.0003$ compared to $0.9138 \pm 0.0007$ for L2C-TabPFN. Similarly, Frog yields a BCA of $0.8056 \pm 0.0019$, outperforming L2C-TabPFN’s $0.7839 \pm 0.0028$. Both improvements are statistically significant ($p = 0.0312$), indicating that Frog offers a clear advantage for diagnostic prediction.

\textbf{ADAS-Cog score prediction:} When forecasting ADAS-Cog scores, lower MAE (mean absolute error) indicates better performance. Here, Frog demonstrates superior accuracy, achieving an MAE of $6.0030 \pm 0.0229$, which is substantially lower than L2C-TabPFN’s MAE of $6.7932 \pm 0.0473$. This improvement is also statistically significant ($p = 0.0312$), further supporting Frog’s effectiveness for this cognitive outcome. However, it is important to note that even in the original TADPOLE follow-up paper~\citep{marinescu2020alzheimer}, ADAS-Cog forecasting performance was poor across all participating teams—worse than simply predicting the median for all subjects. As such, ADAS results should be interpreted with caution, and strong performance on this metric may not be especially meaningful, given the general difficulty of the task.

\textbf{Ventricles volume prediction:} For the Ventricles task, L2C-TabPFN outperforms Frog, yielding a lower MAE of $0.1577 \pm 0.0005$ compared to Frog’s $0.1781 \pm 0.0008$. Again, this difference is statistically significant ($p = 0.0312$), demonstrating that L2C-TabPFN is more effective for ventricular volume forecasting.

Overall, the relative forecasting performance of L2C-TabPFN and Frog is task-dependent. Frog provides significant improvements for diagnostic and cognitive outcomes, while L2C-TabPFN excels at predicting ventricular volume. Each model thus shows a statistically significant advantage for at least one prediction task.

\subsection{Interpretability}

To understand how each model makes its predictions, we used SHAP (SHapley Additive exPlanations)~\citep{lundberg2017unified} values to measure the impact of individual features on model outputs. Figure~\ref{fig:shap_summary} shows summary plots of SHAP values for the ventricles, diagnosis (DX), and ADAS models. Each point on the plots represents the impact of a feature on a single sample, with color indicating the feature value (blue for low and pink for high).

\subsubsection{L2C-TabPFN}

For the ventricles model, the most important features are direct measurements related to ventricular volume, including \textit{mr\_Ventricles}, \textit{high\_Ventricles}, and \textit{low\_Ventricles}. Other brain volumetric features, such as \textit{high\_WholeBrain}, \textit{mr\_WholeBrain}, and measures of intracranial volume (ICV), also contribute, but to a lesser extent. Demographic and genetic features, such as \textit{current\_age}, \textit{APOE4} status, and \textit{is\_male}, have relatively minor effects.

In the DX model, features directly related to clinical diagnosis and cognitive status are most important. The top features include \textit{mr\_DX}, \textit{high\_CDRSB}, \textit{best\_DX}, and \textit{worst\_DX}, all of which reflect clinical or cognitive assessments. Some anatomical features, such as ventricular and hippocampal measurements, also influence predictions, but their impact is smaller compared to clinical variables.

For the ADAS model, there is a broader mix of important features. Both anatomical measurements (such as \textit{low\_Ventricles}, \textit{high\_ICV}, and \textit{mr\_ICV}) and clinical scores (\textit{low\_CDRSB}, \textit{high\_CDRSB}, and \textit{mr\_CDRSB}) contribute to the model's predictions. Demographic factors, such as current age (\textit{current\_age}) and education level (\textit{educ}), as well as genetic factors like \textit{APOE4}, also play a role. This suggests that the ADAS model uses a more varied set of inputs, relying on both anatomical and clinical data.

\subsubsection{Frog (L2C-XGBoost)}

The interpretability of the Frog (L2C-XGBoost) model is assessed similarly using SHAP values. For the ventricles model, SHAP summary plots highlight that the most important features are again direct measurements of ventricular volume, such as \textit{mr\_Ventricles}, \textit{high\_Ventricles}, and \textit{low\_Ventricles}. However, compared to the L2C-TabPFN model, the XGBoost-based model assigns a larger proportion of the total importance to these top features, resulting in a sparser and more focused importance profile. This observation can probably explain the poor performance of the Frog. Volumetric features related to the whole brain and intracranial volume also appear relevant, though their SHAP values are typically much lower than those for ventricular measurements in L2C-TabPFN.

In the DX model, the most influential features identified by XGBoost closely align with a clinical diagnosis and cognitive status, including \textit{mr\_DX}, \textit{high\_CDRSB}, \textit{best\_DX}, and \textit{worst\_DX}. Anatomical variables, such as ventricular and hippocampal volumes, are present but contribute less to the model compared to clinical assessment features, consistent with findings in the L2C-TabPFN model.

For the ADAS model, the Frog model highlights both anatomical measurements and cognitive scores, such as \textit{low\_Ventricles}, \textit{high\_ICV}, \textit{mr\_ICV}, \textit{low\_CDRSB}, and \textit{high\_CDRSB}, as top features. Notably, demographic and genetic features, including \textit{current\_age}, \textit{educ}, and \textit{APOE4}, also contribute but have a more distinct separation in importance compared to the TabPFN model.

\subsubsection{Comparative Interpretability}

While both L2C-TabPFN and Frog (L2C-XGBoost) leverage a core set of clinically and biologically relevant features in their predictions, their approaches to interpretability differ in notable ways. The Frog model (L2C-XGBoost) produces more concentrated and sparser SHAP value profiles, typically focusing on a small number of key features with higher attribution scores. This enables the straightforward identification of primary drivers for each prediction and closely aligns with established clinical knowledge, facilitating transparency and trust in clinical settings.

On the other hand, L2C-TabPFN tends to distribute feature importance more broadly across a larger set of variables. While this can make interpretation less direct, it may also reflect the model's strength in integrating more complex and subtle patterns across diverse data sources.
This distinction is also reflected in model performance: L2C-TabPFN achieves superior accuracy for ventricles prediction, where a broader integration of features may be beneficial, while the Frog model performs better on diagnosis and ADAS-Cog tasks.

\section{Conclusions}

In this work, we proposed L2C-TabPFN and evaluated its performance against Frog, a state-of-the-art XGBoost baseline, on the TADPOLE Challenge dataset. The results demonstrate that while both models achieve high diagnostic accuracy, L2C-TabPFN yields significant improvement in ventricular volume prediction. This improvement suggests that L2C-TabPFN more effectively captures subtle and clinically relevant structural changes in brain imaging data, which are essential for precise monitoring of Alzheimer's disease progression~\citep{marinescu2019tadpole}. These findings highlight the potential of transformer-based models such as TabPFN, particularly when combined with a longitudinal-to-cross-sectional transformation, to advance the state-of-the-art in modeling temporal patient data for regression tasks. In addition to predictive performance, our SHAP-based analysis demonstrates that L2C-TabPFN is capable of identifying clinically relevant features that contribute to its predictions. Future research will extend this evaluation to additional longitudinal datasets to explore their predictive behavior and interpretability in real-world clinical settings.
\bibliography{jmlr-sample}

\appendix

\end{document}